%% file: main.tex
\pdfoutput=1

\documentclass[10pt,twocolumn,letterpaper]{article}

\usepackage[pagenumbers]{cvpr} 

\usepackage{graphicx}
\usepackage{amsmath}
\usepackage{amssymb}
\usepackage{booktabs}
\usepackage{multirow}
\usepackage{bm}
\usepackage{bbding}

\newcommand{\mypara}[1]{\paragraph{#1.}}

\usepackage[pagebackref,breaklinks,colorlinks]{hyperref}

\usepackage[capitalize]{cleveref}
\crefname{section}{Sec.}{Secs.}
\Crefname{section}{Section}{Sections}
\Crefname{table}{Table}{Tables}
\crefname{table}{Tab.}{Tabs.}

\begin{document}

\title{Video-Text Pre-training with Learned Regions}

\author{Rui Yan\thanks{This work is done by Rui Yan when he visited National University of Singapore. Email: ruiyan@njust.edu.cn}~~$^{1,2}$ \quad Mike Zheng Shou$^2$ \quad Yixiao Ge$^{3}\;$
\quad Alex Jinpeng Wang$^2$ \\
\quad Xudong Lin$^4$ 
\quad  Guanyu Cai$^5$ \quad Jinhui Tang\thanks{Corresponding author}~~$^{1}$  \\ \\
$^1$Nanjing University of Science and Technology \quad $^2$Show Lab, National University of Singapore\quad \\ $^3$Applied Research Center~(ARC), Tencent PCG \quad $^4$Columbia University \quad $^5$Tongji University\\}

\maketitle

\begin{abstract}
    Video-Text pre-training aims at learning transferable representations from large-scale video-text pairs via aligning the semantics between visual and textual information. State-of-the-art approaches extract visual features from raw pixels in an end-to-end fashion. However, these methods operate at frame-level directly and thus overlook the spatio-temporal structure of objects in video, which yet has a strong synergy with nouns in textual descriptions. In this work, we propose a simple yet effective module for video-text representation learning, namely \textbf{RegionLearner}, which can take into account the structure of objects during pre-training on large-scale video-text pairs. Given a video, our module (1) first quantizes visual features into semantic clusters, then (2) generates learnable masks and uses them to aggregate the features belonging to the same semantic region, and finally (3) models the interactions between different aggregated regions. In contrast to using off-the-shelf object detectors, our proposed module does not require explicit supervision and is much more computationally efficient. We pre-train the proposed approach on the public WebVid2M and CC3M datasets. Extensive evaluations on four downstream video-text retrieval benchmarks clearly demonstrate the effectiveness of our RegionLearner. The code will be 
   available at \url{https://github.com/ruiyan1995/Region_Learner}.

\end{abstract}

\input{1_intro}

\input{2_related_work}

\input{3_approach}

\input{4_exp}

\input{5_conclusion}

\section*{Acknowledgement}
\noindent Rui Yan is supported by the China Scholarship Council under Grant 202006840101. This work is supported by the National Research Foundation of Singapore under its NRFF award NRF-NRFF13-2021-0008. Mike Zheng Shou received funding only from National Research Foundation, Singapore.
We would like to thank Yuying Ge for her kindly help on distributed training.

{

\input{main.bbl}
}

\end{document}

%% file: 1_intro.tex
\input{motivation}

\section{Introduction}
Video-Text pre-training~\cite{sun2019videobert,miech2019howto100m,liu2019use,liu2021hit,lei2021less,bain2021frozen}, which aims to learn transferable representations by aligning the semantics of video and text, has attracted researchers' attention in recent years. It enables a series of downstream video-text tasks, such as video-text retrieval~\cite{rohrbach2015dataset,krishna2017dense,xu2016msr}, video question answering~\cite{jang2017tgif,lei2018tvqa,lei2019tvqa+}, and video captioning~\cite{zhou2018towards,rohrbach2015dataset,xu2016msr}. The conventional pipeline~\cite{bain2021frozen,lei2021less,sun2019videobert,zhu2020actbert} of video-language pre-training is encoding video and text into the shared feature space followed by the cross-modality modelling. The visual input used in existing methods can be categorized as \emph{whole frames} and \emph{whole frames + explicit object boxes}, as shown in Figure~\ref{fig:mov_a}.


 \textbf{i), whole frames}~\cite{sun2019videobert,lei2021less,bain2021frozen}: are directly encoded as the video features through the pre-trained 2D or 3D visual backbone. Limited to computing resources, early works~\cite{liu2019use,sun2019videobert} extract such video features in an offline way, but recent methods have managed to train the visual backbone on the raw frames in an end-to-end manner~\cite{bain2021frozen,lei2021less}. Whereas, these methods evenly encode each frame as a number of patch-features, which inevitably destroys the inherent spatio-temporal structure of the visual entities.

\textbf{ii), whole frames + explicit object box}~\cite{zhu2020actbert,liu2019use}: extract semantic region features from explicit object boxes detected by off-the-shelf algorithms~\cite{renNIPS15fasterrcnn,redmon2016you} for better performance. Intuitively, the visual information of local objects is more effective for semantic alignment. However, existing methods adopt offline region features which are very computationally expensive and not 
flexible. Beyond that, region features used in these methods also heavily rely on the quality of the off-the-shelf detectors.

{

}

{
These observations motivate us to design lightweight approach to \textbf{implicitly learning object region~(as shown in Figure~\ref{fig:mov_a}~(iii))} without position supervision. We propose a simple yet effective plug-and-play RegionLearner module for video-text representation learning. 
Specifically, this module is performed via the following three steps. \textbf{i) Quantization:} To construct possible region areas, we quantize visual features 
learned from raw pixels into semantic cluster representations.
\textbf{ii) Aggregation}: aggregate them in each semantic regions/objects through multiple learnable region masks; \textbf{iii) Interaction:} based on these aggregated features, we can easily build spatio-temporal interactions between different visual entities for a better understanding of video data. It is noteworthy that our method does not require explicit supervision and is computationally efficient. 


}

Our contributions can be summarized as three-fold:
\begin{itemize}
    \item To our best knowledge, we are the first to take into account the structure of objects in video during video-text pre-training in an end-to-end manner. We propose a framework to implicitly learn discriminative regions from patch-features without explicit supervision.
    \item \textbf{RegionLearner} is proposed to quantize raw visual pixels into discrete latent features, and aggregate them with each semantic region for further interaction. It is compatible with different encoders or learning objectives.
    \item We have validated the proposed method on four downstream benchmarks, and the results significantly outperform the existing methods. In particular, our approach gains absolute improvement of {$5.3\%$ \textbf{R@1} score on MSR-VTT~\cite{xu2016msr} 1K-A and $10.3\%$ \textbf{R@1} score on MSVD~\cite{chen2011collecting}}. We will release relevant code and pre-trained model weights to facilitate the research community.
\end{itemize}




%% file: motivation.tex

\begin{figure}[t]
     \centering
     \begin{subfigure}[b]{0.47\textwidth}
         \centering
         \includegraphics[width=\textwidth]{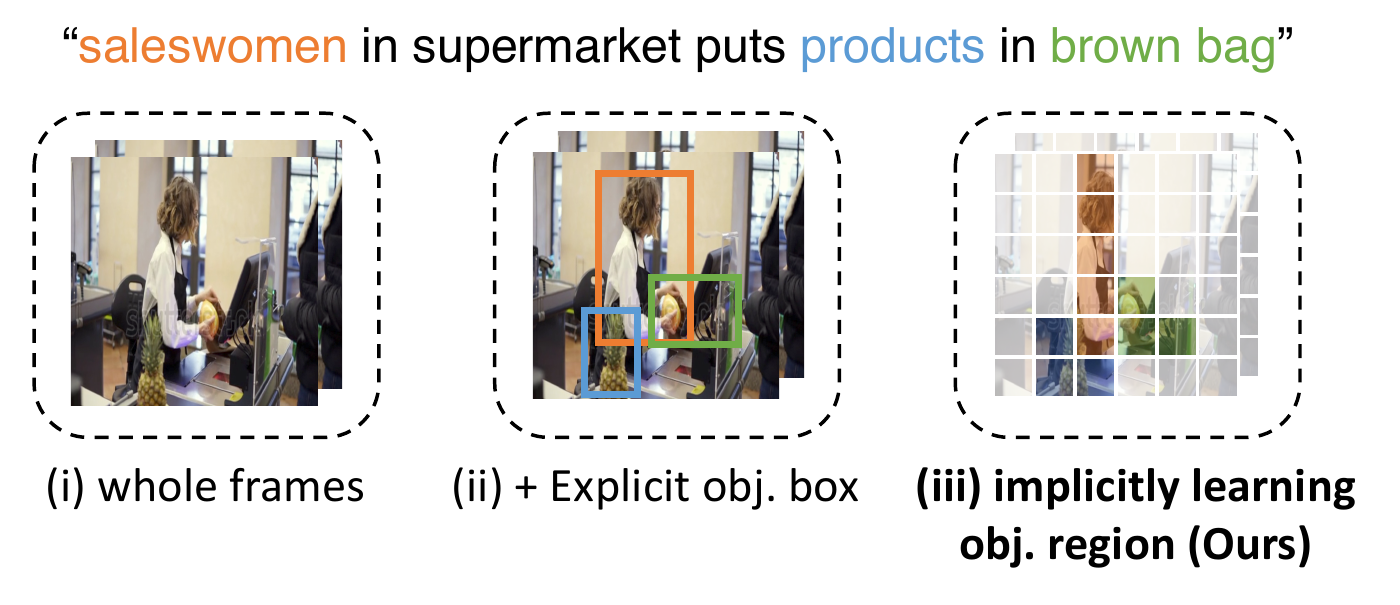}
         \caption{Previous works for video-text pre-training usually extract visual features from the whole frames of video. Inspired by the success of region features in image-text representation~\cite{li2020oscar,chen2020uniter,lu2019vilbert,Su2020VL-BERT}, some video-based works also extract semantic features~\cite{zhu2020actbert,liu2019use} from explicit object regions. Our motivation is to implicitly learn object regions from raw pixels without any supervision. 
         }
         \label{fig:mov_a}
     \end{subfigure}
     \hfill
     \begin{subfigure}[b]{0.47\textwidth}
         \centering
         \includegraphics[width=\textwidth]{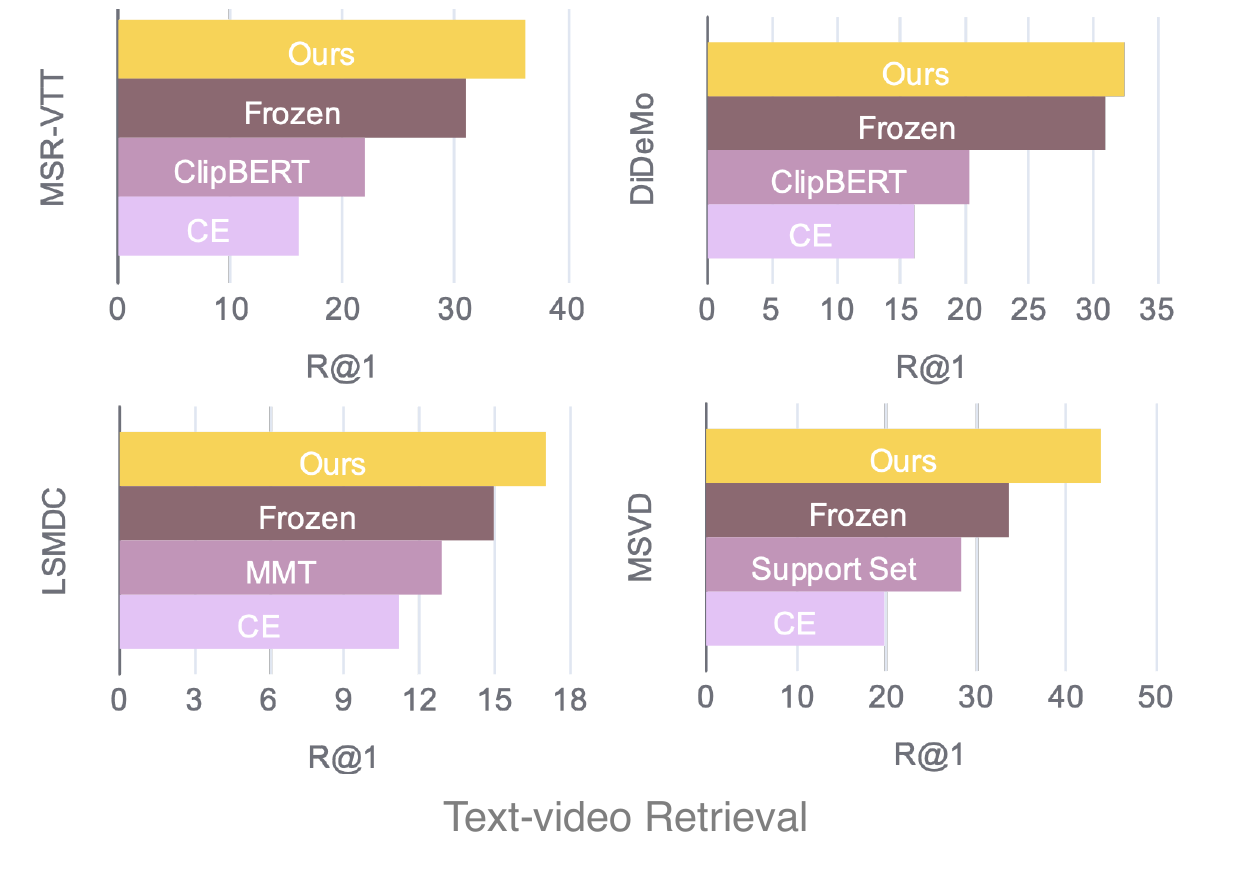}
         \caption{By learning implicitly object regions from grid features, our approach brings significant improvements on four text-video retrieval benchmarks.}
         \label{fig:mov_b}
     \end{subfigure}
        \caption{Our main motivation and results.}
        \label{fig:motivation}
\end{figure}

%% file: 2_related_work.tex
\section{Related Work}

\mypara{Vision-Language Pre-training}
{Learning visual representation from large-scale video-text pair collections is an emerging research topic. Early methods~\cite{kay2017kinetics,sun2019videobert,li2020hero} extract offline visual features from pre-trained video backbones for pre-training. Some recent methods~\cite{lei2021less,bain2021frozen} directly extract visual features from raw pixels in an end-to-end fashion. For example, ClipBERT~\cite{lei2021less} learns joint representations directly from pixels of sparse sampled frames and raw text tokens. Besides, some works~\cite{liu2019use,zhu2020actbert} attempt to extract regional features from videos as supplementary with the help of off-the-shelf detectors~\cite{anderson2018bottom} pre-trained on Visual Genome~\cite{krishna2017visual}. For instance, ActBERT~\cite{zhu2020actbert} mines the fine-grained visual clues from object regions, including regional object features and the position of objects. However, frame feature~\cite{lei2021less,bain2021frozen} used in existing video-language pre-training methods ignore the complete semantics of visual objects, meanwhile region features heavily rely on the quality of detectors. In this work, we implicitly learn regions from raw pixels without object boxes for video-language pre-training in an end-to-end fashion.


Some recent works for image-text pre-training attempt to get rid of the regional feature~\cite{tan2019lxmert,chen2020uniter,kim2018bilinear} which has been dominant in image-text representations. They either randomly sample some patch-features~\cite{huang2020pixel} or construct compact discrete representations through visual dictionary~(clustering)~\cite{huang2021seeing} to achieve promising performance. Because semantics involved in language descriptions are visually intertwined, simply clustering~\cite{huang2021seeing} or randomly sampling~\cite{huang2020pixel} patch-features will inevitably lead to a large number of fragmented and incomplete areas. In this work, we not only quantizes visual feature into semantic clusters inspired by~\cite{huang2021seeing}, but also further aggregate discrete representation belonging to the same semantic region followed by interactions.

}

\input{overview}

\mypara{Video Representation Learning}
In the past decade, a large number of deep 2D~\cite{karpathy2014large,simonyan2014two, lin2019tsm,wang2016temporal,feichtenhofer2016convolutional} and 3D~\cite{carreira2017quo, tran2015learning, wang2018non, feichtenhofer2019slowfast, hara2018can, zhang2021morphmlp} models have been proposed to extract efficient spatial and temporal representations for video. Recently, inspired by the success of Transformer in NLP field~\cite{vaswani2017attention,devlin2018bert}, visual Transformers~\cite{liu2021video,cheng2021motion,neimark2021video,arnab2021vivit} are sprung up for video representation. However, these pre-trained video backbones focus more on temporal cues of defined action categories~\cite{kay2017kinetics,sigurdsson2016hollywood} from the whole frames, cannot cover rich spatial semantics involved in language descriptions. In this work, we adopt the video backbone pre-trained from ImageNet~\cite{deng2009imagenet}, and then learn more clues from local visual entities.

To model the dynamic motion of objects, some recent works~\cite{yan2020interactive,gao2018ican,materzynska2020something,herzig2021object} extract the instance-centric features from a video according to the ground-truth or detected tracklets. Different from these instance-centric works which need explicit positional supervision of regions, we aim at implicitly learning object regions from the raw frames directly. 





%% file: overview.tex
\begin{figure*}[t]
  \centering
   \includegraphics[width=0.97\linewidth]{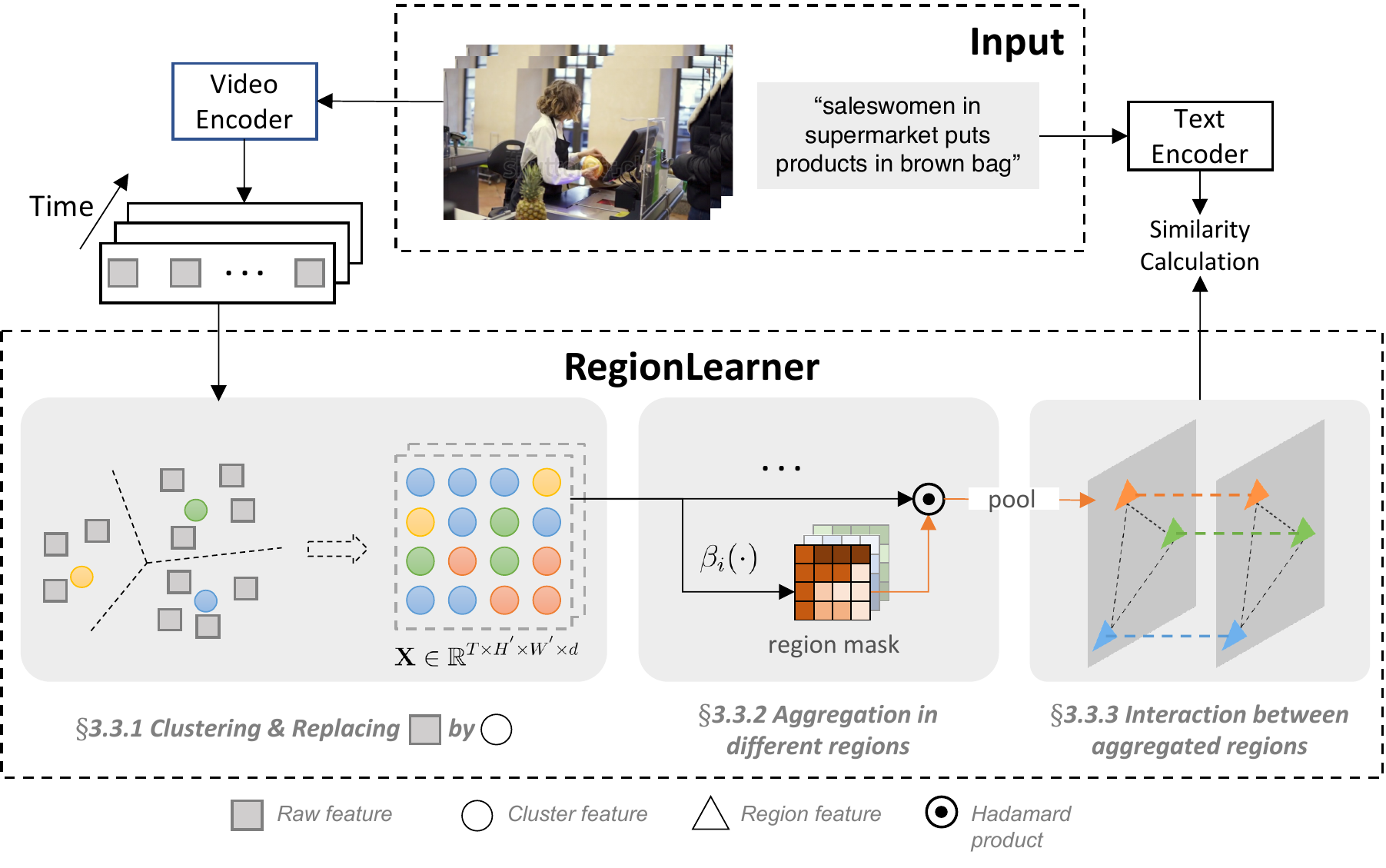}
   \caption{Overview of the proposed approach. Given a video-text pair, we encode them via the video and text encoder respectively. Region Learner is proposed to identify and leverage the implicit semantic structure of objects/regions in video, and it is performed in three steps. \textbf{i) Quantization~(Sec.~\ref{sec::RL_con}):} clustering the raw features into semantic clusters and replace raw features with cluster features;
   \textbf{ii) Aggregation~(Sec.~\ref{sec::RL_agg}):} we design multiple learnable masks to aggregate information with each semantic region; \textbf{iii) Interaction~(Sec.~\ref{sec::RL_int}):} perform spatio-temporal interactions between different aggregated regions. Finally, the video representation generated from Region Learner is used to compute the similarity with textual representation.}
   \label{fig:overview}
\end{figure*}


%% file: 3_approach.tex
\section{Approach}
{In this section, we present the main pipeline of our approach in Section~\ref{Overall} and the details in Section~\ref{sec::encoder} and~\ref{sec::Region_Learner}. The objective function used in pre-training is described in Section~\ref{sec::obj_fun}}

\subsection{Overall}~\label{Overall}
We show the overall architecture of the proposed video-language pre-training framework in Figure~\ref{fig:overview}. It consists of two trainable visual and text encoders and the proposed {RegionLearner}. Given a pair of video-text, two encoders extract visual and textual features respectively~\ref{sec::encoder}. {RegionLearner}~\ref{sec::Region_Learner} is designed to quantize raw pixels into discrete latent representations and aggregate them 
into different object regions via learnable region masks. Before computing the similarity of these features from two modalities, we further explore the spatio-temporal relations among these semantic region features.

\subsection{Video and Text Encoder}\label{sec::encoder}
Previous works usually extract offline visual feature from large-scale pre-training datasets limited to the computation before cross modality modeling. Recently, several works visual feature extracted from raw video or image in an end-to-end fashion shows strong discrimination. In this work, we extract both visual and textual features in a trainable way, similar to~\cite{bain2021frozen,lei2021less,huang2021seeing}. 

Formally, supposing the input of our approach is a video $\bm V \in \mathbb{R}^{T \times 3 \times H \times W}$ which contains $T$ frames of resolution $H \times W$, and the associated tokenized textual description $\bm C$. We obtain their features as,
\begin{equation}
    {\bm F} = {E}_{\mathrm V}(\bm V), {\bm Y} = {E}_{\mathrm C}(\bm C).
\end{equation}Here, ${E}_{\mathrm V}(\cdot)$ and ${E}_{\mathrm C}(\cdot)$ are video and text encoder, respectively. For the text encoder, we choose the current most popular transformer-based structure and treat the [CLS] token of the last hidden layer as the text feature ${\bm Y}$. Either convolution or transformer can be used as a video encoder. Here, we extract visual representations from a video~$\bm V$ via ViT~\cite{dosovitskiy2020image,bertasius2021space} with the patch size of $P$, thus we obtain video feature ${\bm F} \in \mathbb{R}^{T \times L \times d}$ where $L=HW/P^{2}$. 

\subsection{RegionLearner}\label{sec::Region_Learner}
In this section, we present the details of the proposed RegionLearner. It consists of the following three steps, quantify raw continuous video features into discrete representation~(\textbf{Quantization}, Sec.~\ref{sec::RL_con}), and aggregate them into regions~(\textbf{Aggregation}, Sec.~\ref{sec::RL_agg}), and finally build relationships between each of them~(\textbf{Interaction}, Sec.~\ref{sec::RL_int}).

\subsubsection{Quantization}\label{sec::RL_con}


{
\textbf{Motivation.}~While language used by humans is inherently discrete, the visual information from video is continuous and diverse. Trying to align concepts from these two modalities is already difficult, let alone without any auxiliary supervision (e.g., semantic object annotation). Therefore, in such a situation, we believe that quantifying raw continuous visual features to discrete representations will make video-language alignment easier. In the past, some classic feature aggregation methods (\emph{e.g.}, VLAD~\cite{jegou2010aggregating}, NetVLAD~\cite{arandjelovic2016netvlad}) were used to construct a fixed number of feature clusters to quantify image representation, and they have been proven effective in both image and video representation~\cite{huang2021seeing}. Inspired by this, we cluster the similar local visual features extracted from patches of each frame in the global space, as follows.

\noindent\textbf{Implementation.}~Formally, we first define $M$ learnable clusters as $\{\bm c_{0}, \bm c_{1}, \cdots, \bm c_{M}\}$ in which $\bm c_{m} \in \mathbb{R}^{d^{'}}$. Our main idea is to aggregate similar visual tokens into the shared cluster in the global space~(over the entire dataset). Given a $f_{s}^{t}$ at the $s$-th spatial position and $t$-th time-step, we update it with the most similar cluster as,
\begin{align}
    {f_{s}^{t}}^{'} &= {\bm c_{m^{*}}},\quad \mathrm{where} \quad m^{*} = \mathrm{argmin}_{m}~\mathrm{dis}(f_{s}^{t}, \bm c_{m}).
\end{align}Here $\mathrm{dis}(a, b)$ is used to compute the similarity distance between two input features, and it can be implemented by different methods~({\em e.g.}, euclidean distance, and cosine similarity). In this work, we adopt euclidean distance and update these clusters with momentum learning and stop the gradient on the operation of $\mathrm{argmin}$ following~\cite{oord2017neural,huang2021seeing}. After that, we can achieve a more compact representation $\bm F^{'} \in \mathbb{R}^{T \times L \times d}$ with the same shape of input feature $\bm F$.

}

\subsubsection{Aggregation}\label{sec::RL_agg}
{
\textbf{Motivation.}~According to visual similarity, video features are quantized through a limited number of clusters, making the local feature representation more compact. But it is inevitable to fragment the semantics of visual instances. In other words, the model tends to represent a complete visual instance by combining several different cluster features. Intuitively, the semantic unit of language usually refers to a visual instance, or region. Therefore, it is necessary to further abstract several visual representations with complete semantics from each quantized feature map. Specifically, inspired by~\cite{ryoo2021tokenlearner}, we extract $K$ region features with complete semantics in space from the sequential features~$\bm F^{'}$.

}
%


{
\noindent\textbf{Implementation.}~Formally, for each video, we can obtain the semantic representation $\bm F^{'} \in \mathbb{R}^{T \times L \times d}$ in which patch features are arranged in sequence. To find semantic regions, we reshape it back to the original spatial resolution $\bm X \in \mathbb{R}^{T \times H^{'} \times W^{'} \times d}$ where $H^{'}=H/P$ and $W^{'}=W/P$. For each frame $t$, we aim at learning $K$ region representations $\bm S_{t}=[\bm s_{i}]^{K}_{i=1}$ from input frame feature via:
\begin{equation}
    \bm s_{i} = \mathcal{R}(\bm X_{t}),
\end{equation}where $X^{'}_{t} \in \mathbb{R}^{H^{'} \times W^{'} \times d}$ and ${\bm s}_{i} \in \mathbb{R}^{d}$. This way, region features aggregate informative pixels~({\em i.e.}, small patches from a video frame) adaptively. 

Notably, $\mathcal{R}(\cdot)$ can be implemented via different choices. In this work, we instantiate this function as multiplying the input feature map $\bm X_{t}$ by a learned spatial attention map and pooling it to a single vector,
\begin{equation}
    \bm s_{i} = \mathcal{R}(\bm X_{t}) = \mathtt{Pool2D}(\bm \beta_{i}(\bm X_{t}) \circ \bm X_{t}).
\end{equation}Here, $\mathtt{Pool2D}$ and $\circ$ denote the spatial pooling and Hadamard product respectively. $\bm \beta=[\bm \beta_{i}]_{i=1}^{K}$ is implemented by a $3 \times 3$ convolution layer with $K$ channels.  After that, for each video, sparse region features $\bm S \in \mathbb{R}^{S \times T \times d}$ are obtained, which allows us to further model the spatio-temporal clues in video.

 
}

\subsubsection{Interaction}\label{sec::RL_int}
{

\textbf{Motivation.}~Video descriptions will inevitably contain dynamic motions, such as ``put sth." and ``pull sth.", which involves the temporal and spatial dynamic relationships between visual entities rather than only static appearance. Therefore, we further build spatio-temporal interactions among the region features, which is more efficient and flexible than modeling at the pixel level. 

\noindent\textbf{Implementation.}~The small number of region features allows us to directly build space-time attention as,
\begin{equation}
    \bm z_{k,t} = \sum_{k^{'},t^{'}} \bm \alpha(\bm \phi(\bm s_{k,t}), \bm \phi(\bm s_{k^{'},t^{'}}))\bm s_{k^{'},t^{'}},
\end{equation}where $\bm \phi(\cdot)$ is linear embedding, $\bm \alpha(\bm a, \bm b)$ computes the attention weights between input via dot product followed by a softmax~\cite{bertasius2021space}. $\bm Z \in \mathbb{R}^{K \times T \times d}$ is the video feature output from RegionLearner.
}

\subsection{Objective function}\label{sec::obj_fun}
{Following~\cite{bain2021frozen}, in each training batch, paired video-text samples are treated as positives and others are negatives. Formally, we minimise the sum of video-to-text loss~($\mathcal{L}_{\mathrm{v2t}}$) and text-to-video loss~($\mathcal{L}_{\mathrm{t2v}}$) as follows:
\begin{align}
    \mathcal{L}_{\mathrm{v2t}}=-
    \frac{1}{N}\sum_{i=1}^{N}\mathrm{log}\frac{\mathrm{exp}({\bm z}_{i}^{\top}{\bm y}_{j}/\tau)}{\sum_{j}^{N}{\mathrm{exp}({\bm z}_{i}^{\top}{\bm y}_{j}/\tau)}},\\
    \mathcal{L}_{\mathrm{t2v}} = -
    \frac{1}{N}\sum_{i=1}^{N}\mathrm{log}\frac{\mathrm{exp}({\bm y}_{i}^{\top}{\bm z}_{j}/\tau)}{\sum_{j}^{N}{\mathrm{exp}({\bm y}_{i}^{\top}{\bm z}_{j}/\tau)}},
\end{align}where ${\bm z}_{i}$ and ${\bm y}_{i}$ are normalized video and text features, respectively. $N$ is the batch-size, and temperature variable $\tau$ is used to scale logits.

}

%% file: 4_exp.tex
\section{Experiments}
{In this section, we first introduce pre-training and downstream datasets in Section~\ref{sec::Pretraining_Datasets} and \ref{sec::Downstream_Datasets}. Implementation details of the proposed model and the training details are present in Section~\ref{sec::Implementation_Details}. After that, some ablation studies of our approach on MSR-VTT~\cite{xu2016msr} are shown in Section~\ref{sec::Ablation}. Finally, we compare our approach with the state-of-the-art methods on four downstream benchmarks in Section~\ref{sec::re_VTR}.


}

\subsection{Pre-training Datasets}\label{sec::Pretraining_Datasets}
{Training the model in an end-to-end way on large video datasets~({\em{e.g.}}, HowTo100M~\cite{miech2019howto100m}) is impractical. Following the recent work~\cite{bain2021frozen}, we pre-train our model on an affordably large-scale video-text benchmark~(WebVid-2M~\cite{bain2021frozen}) and an image-text benchmark~(Google Conceptual Captions~\cite{sharma2018conceptual}).

\noindent\textbf{WebVid2M~\cite{bain2021frozen}}~contains about $2.5$M image-text pairs harvested from the web. Most pairs have visual and language well-aligned, thanks to manually generated and well-formed captions. However, there are limited temporal dynamics involved in this benchmark, although each video covers an average of $18$ seconds. 


\noindent\textbf{Google Conceptual Captions~(CC3M)~\cite{sharma2018conceptual}}~contains about $3.3$M image-text pairs. These raw text descriptions are gleaned from HTML attributes associated with the image. CC-3M exhibits large diversity in style.



}

\subsection{Downstream Benchmarks}\label{sec::Downstream_Datasets}
{In this work, we evaluate the effectiveness of our pre-trained model on four downstream benchmarks as follows.}

\noindent\textbf{MSR-VTT~\cite{xu2016msr}}~provides $10$K video clips with about $200$K video-text pairs, It is the most popular video-text retrieval benchmark including most of the categories and visual content in the real world. Following previous works~\cite{liu2019use,bain2021frozen}, we adopt the split of $9$K and $1$K videos for training and testing.

\noindent\textbf{DiDeMo~\cite{anne2017localizing}}~consists of $10$K videos each of which may labeled with multiple sentences, leading in total $40$K sentences. In our experiments, we treat all sentences of each video as a single description following~\cite{liu2019use,lei2021less,bain2021frozen}. Note that, we do not use the ground-truth localization annotations of this dataset.

\noindent\textbf{LSMDC~\cite{rohrbach2015dataset,rohrbach2017movie}} used in this work is LSMDC-2016 which consisting of in total $128$K clips. For a fair comparison, we follow the setting in~\cite{bain2021frozen,rohrbach2017movie} to validate our approach on the test set of $1,000$ video clips.

\noindent\textbf{MSVD~\cite{chen2011collecting}}~is a relative small benchmark consisting of only $1,970$ videos, each of which is described by $40$ English sentences. We split these videos into $1200$, $100$, and $670$ as the train, val, and test set~\cite{liu2019use,bain2021frozen,patrick2020support}.


\subsection{Implementation Details}\label{sec::Implementation_Details}
{
\noindent\textbf{Input.}~Following~\cite{bain2021frozen}, the number of input frames for one video used in our model is (no more than) $4$ during pre-training while $8$ during fine-tuning on downstream benchmarks. Each video is resized to $224 \times 224$ followed by some video-based augmentations~(horizontal flip and color jitter).

\noindent\textbf{Model.}~Video and text encoders are respectively instantiated with TimeSformer~\cite{bertasius2021space} pretrained on ImageNet and DistilBERT base-uncased~\cite{sanh2019distilbert} pretrained on BookCorpus~\cite{zhu2015aligning} and English Wikipedia. The outputs of text encoder are projected into a $256$-dimension vector by a linear layer. Video encoder divides each frame into $L=196$ patches with the patch-size $P=16$, and then embeds these patches into a $768$-dimension vector followed by $12$ attention blocks with $12$ heads.

\noindent\textbf{Training.} We implement the approach with PyTorch on Linux and train all models on Tesla A100 GPUs with the batch-size of $256$. We use the Adam optimizer with different learning rate schedule on pre-training and downstream training. For pre-training, we train our approach in $50$ epochs using an initial learning rate of $2 \times e^{-4}$. The learning rate decays to $1/10$ of the previous one at $30$ and $40$ epochs. Besides, only $1$M video-text pairs are sampled for each epoch. For fine-tuning, all experiments on downstream benchmarks are trained in $100$ epochs with $T=8$ frames and a fixed $3 \times e^{-5}$ learning rate. The whole pre-training duration is about $2$ days on $16$ GPUs. Besides, the training on downstream benchmarks takes no more than $2$ hours on $4$ GPUs.

}

\input{MSRVTT}

\subsection{Comparisons with State-of-the-art}\label{sec::re_VTR}
{We compare the proposed approach with the state-of-the-art methods on four benchmarks, as shown in Table~\ref{tab:MSRVTT},~\ref{tab:DiDeMo},~\ref{tab:LSMDC},~\ref{tab:MSVD}. Our approach achieves new SOTA results on all of these four datasets by a significant margin.
}

\noindent\textbf{Results on MSR-VTT.}
{Baseline methods can be categorized as \textbf{i)}, no-pretraining~(\emph{i.e.,}~\cite{yu2018joint,liu2019use,patrick2020support}); \textbf{ii)}, pretrained on HowTo100M~(\emph{e.g.,}~\cite{miech2019howto100m,zhu2020actbert,li2020hero}); \textbf{iii)}, pretrained on relative small benchmarks~(\emph{i.e.,}~\cite{lei2021less,portillo2021straightforward,radford2021learning,bain2021frozen}). 
Our approach exceeds the best no-pretrained method, Support Set~\cite{patrick2020support}, by $8.9\%$. \textbf{In addition, we are superior to all previous methods pre-trained on HowTo100M~\cite{miech2019howto100m} that is an order of magnitude larger than WebVid2M~\cite{bain2021frozen} + CC3M~\cite{sharma2018conceptual}.} Though some of these methods adopt expert features including object~\cite{zhu2020actbert}, sound~\cite{gabeur2020multi,rouditchenko2020avlnet} and speech~\cite{gabeur2020multi} information. Compared with the most related work, Frozen~\cite{bain2021frozen}, our {RegionLearner} brings significant improvements on text-to-video retrieval. We also provide the results of the zero-shot setting which requires models not to fine-tune on the downstream benchmark. Our proposed {RegionLearner} boosts Frozen by $3.5$ on {R@1} in Text-Video retrieval and achieves state-of-the-art results against other methods.



}

\input{DiDeMo}

\noindent\textbf{Results on DiDeMo.}
{As expected, our approach outperforms all existing results on this benchmark. Compared with the most related work~(Frozen~\cite{bain2021frozen}), our approach gains a boost of $1.5\%$ on {R@1}, which is not very significant. Because this benchmark is collected for Moments Localization~\cite{anne2017localizing}, the provided text caption describes only part of the video. Our {RegionLearner} will be disturbed by many irrelevant frames, which may be an interesting problem for future research.


}

\noindent\textbf{Results on LSMDC.}
{We also report the text-to-video retrieval result on LSMDC in Table~\ref{tab:LSMDC}. Our approach surpassed all existing methods reported on this benchmark and improves the existing state-of-the-art Frozen by approximately $2.0\%$ in terms of R@1 and MedR. 
}

\input{LSMDC}

\noindent\textbf{Results on MSVD.}{
Despite this benchmark is relatively small, our method is still effective and achieves new state-of-the-art results, as shown in Table~\ref{tab:MSVD}. In particular, compared with the existing best result from Frozen, our approach improves R@1 by $10.3\%$. It suggests that our method is still effective on the small-scale downstream dataset, which is very important for pre-training methods to be employed to deal with various real-world tasks. 

}

\input{MSVD}


\subsection{Ablation Study}\label{sec::Ablation}
In this section, we study the effectiveness of each component of the proposed approach and the effect of different parameters used in model architecture. All the following experiments are pre-trained on WebVid-2M~\cite{bain2021frozen} and fine-tuned on the MSR-VTT~\cite{xu2016msr} and the results of the $1$K-A test set are reported.

\input{component}
\noindent\textbf{Effectiveness of Each Component.}
To demonstrate the effectiveness of the component of the proposed {RegionLearner}, we gradually drop each step used in module and report the results in Table~\ref{tab:component}. In general, each component brings improvements. Among them, the improvement of Quantization and Aggregation is the most obvious, up to $2+\%$. But the gain of Interaction is relatively limited, which may be due to the limited reasoning semantics in pre-training data.

\noindent\textbf{Different Aggregation Strategy.}
{In this work, we compare different aggregation strategies used in {RegionLearner}, such as Random Sampling~\cite{huang2020pixel}, Manual Selection, Naive Attention, and the proposed Region Mask. Random Sampling: samples part of patch feature from $X$ randomly; Manual Selection: selects some high-frequent cluster representations from the map via a threshold; Naive Attention: performs a simple attention mechanism on the feature map. All these methods bring limited improvement, but our region mask improves the base method by approximately $2\%$.

\input{tab6}

}

\noindent\textbf{Number of Regions}
To determine how many regions the model needs to learn, we set the range of $K$ from $2^{0}$ to $2^{6}$, and the results are shown in Figure~\ref{fig:num_region}. We can see that if $K$ is too large, it may be difficult for the model to find discriminative regions because the module tends to reserve the whole feature map. On the contrary, if $K$ is too small many fragmentary and weak semantics will be discarded in large quantities, leading to the poor results. Our approach achieves the best results with $K=8$ regions.

\input{region_depth}

\input{vis}

\noindent\textbf{Depth of Interaction.}
We tried to build multiple layers of interaction on top of the region features and found that too many layers are not good, as shown in Figure~\ref{fig:depth}. Probably because the highly abstract regional features are sufficient, and too many spatio-temporal interactions will cause the deep model to over-fit on the pre-training dataset. Thus, we use only single-layer interaction.


\subsection{Visualization}\label{sec::Visualization}
{We also provide some qualitative results in Figure~\ref{fig:vis} to show the regions learned via the proposed {RegionLearner} on pretraining dataset, WebVid-2M~\cite{bain2021frozen}. 
The second column of each group is the indices map generated by \emph{Quantization}, it represents similar visual patches with the same index. It is worth noting that the different colors in this map only represent different index values. As we can see, \emph{Quantization} is adept at clustering the backgrounds but cannot focus on specific visual entities in the foregrounds. After performing \emph{Aggregation}, we achieve some region masks as shown in the last two columns of each group. Particularly, these region masks not only capture the visual entities in foregrounds but also reserve discriminative background information for alignment. {We also provide two failure cases at the bottom of Figure~\ref{fig:vis}.}

}


%% file: MSRVTT.tex
\begin{table*}[!t]
  \centering
  \begin{tabular}{@{}lc|rrrr|cccc@{}}
    \multirow{2}*{\textbf{Method}}  & \multirow{2}*{\textbf{PT dataset}} &\multicolumn{4}{c}{\textbf{Text} $\Longrightarrow$ \textbf{Video}} &\multicolumn{4}{|c}{\textbf{Video} $\Longrightarrow$ \textbf{Text}}\\
    && \textbf{R@1} & \textbf{R@5} & \textbf{R@10} & \textbf{MedR} & \textbf{R@1} & \textbf{R@5} & \textbf{R@10} & \textbf{MedR}\\
    \hline\hline
    JSFusion~\cite{yu2018joint}  &\multirow{3}{*}{$-$}  &$10.2$ &$31.2$ &$43.2$ &$13.0$ &$-$ &$-$&$-$&$-$\\
    CE~\cite{liu2019use}  &  &$20.9$ &$48.8$ &$62.4$ &$6.0$ &$20.6$ &$50.3$&$64.0$&$5.3$\\
    Support Set~\cite{patrick2020support}  &  &$27.4$ &$56.3$ &$67.7$ &$3.0$ &$26.6$ &$55.1$ &$67.5$ &$3.0$\\
    \hline
    HT MIL-NCE~\cite{miech2019howto100m} & \multirow{11}{*}{HowTo100M~\cite{miech2019howto100m}}     &$14.9$ &$40.2$ &$52.8$ &$9.0$ &$-$ &$-$&$-$&$-$\\
    ActBERT~\cite{zhu2020actbert}  &  &$16.3$ &$42.8$ &$56.9$ &$10.0$ &$-$ &$-$&$-$&$-$\\
    HERO~\cite{li2020hero} & &$16.8$ &$43.4$ &$57.7$ &$-$ &$-$ &$-$&$-$&$-$\\
    UniVL~\cite{luo2020univl}  & &$21.2$ &$49.6$ &$63.1$ &$6.0$ &$-$ &$-$&$-$&$-$\\
    MMT~\cite{gabeur2020multi}   &  &$26.6$ &$57.1$ &$69.6$ &$4.0$ &$27.0$ &$57.5$&$69.7$&$3.7$\\
    Support Set~\cite{patrick2020support}   &  &$30.1$ &$58.5$ &$69.3$ &$3.0$ &$28.5$ &$58.6$&$71.6$&$3.0$\\
    AVLnet~\cite{rouditchenko2020avlnet}  &  &$27.1$ &$55.6$ &$66.6$ &$4.0$ &$-$ &$-$&$-$&$-$\\
    VidTranslate~\cite{korbar2020video}   &   &$14.7$ &$-$ &$52.8$ &$-$ &$-$ &$-$&$-$&$-$\\
    Noise-Estimation~\cite{amrani2021noise}   & &$17.4$ &$41.6$ &$53.6$ &$8.0$ &$-$ &$-$&$-$&$-$\\
    HIT~\cite{liu2021hit}   & &$30.7$ &$60.9$ &$73.2$ &${2.6}$ &$32.1$ &$62.7$&$74.1$&${3.0}$\\
    DECEMBERT~\cite{tang2021decembert}  & &$17.5$ &$44.3$ &$58.6$ &$9.0$ &$-$ &$-$&$-$&$-$\\
    \hline
    ClipBERT~\cite{lei2021less}   & COCO~\cite{chen2015microsoft}, VG~\cite{krishna2017visual} &$22.0$ &$46.8$ &$59.9$ &$6.0$ &$-$ &$-$&$-$&$-$\\
    Frozen~\cite{bain2021frozen}  &CC3M~\cite{sharma2018conceptual}, WV2M~\cite{bain2021frozen} &$31.0$ &$59.5$ &$70.5$ &$3.0$ &$-$ &$-$&$-$&$-$\\
    \textbf{Ours} &CC3M~\cite{sharma2018conceptual}, WV2M~\cite{bain2021frozen} & $\mathbf{36.3}$ &$\mathbf{63.9}$ &$\mathbf{72.5}$ &$\mathbf{3.0}$ &$\mathbf{35.3}$ &$\mathbf{63.5}$&$\mathbf{73.2}$&$\mathbf{3.0}$\\
    \hline\hline
    \textit{Zero-shot} \\
    \hline\hline
    HT MIL-NCE  & HowTo100M~\cite{miech2019howto100m}  &$7.5$ &$21.2$ &$29.6$ &$38.0$ &$-$ &$-$ &$-$ &$-$\\
    Support Set  & HowTo100M~\cite{miech2019howto100m} &$12.7$ &$27.5$ &$36.2$ &$24.0$ &$8.7$ &$23.0$ &$31.1$ &$31.0$\\
    Frozen~\cite{bain2021frozen}   &CC3M~\cite{sharma2018conceptual}, WV2M~\cite{bain2021frozen}  &${18.7}$ &${39.5}$ &${51.6}$ &${10.0}$ &$-$ &$-$ &$-$ &$-$\\
    \textbf{Ours}  &CC3M~\cite{sharma2018conceptual}, WV2M~\cite{bain2021frozen} &$\mathbf{22.2}$ &$\mathbf{43.3}$ &$\mathbf{52.9}$ &$\mathbf{8.0}$ &$\mathbf{15.3}$ &$\mathbf{32.4}$ &$\mathbf{42.1}$ &$\mathbf{17.0}$\\
  \end{tabular}
  \caption{Comparisons with state-of-the-art results on MSR-VTT 1K-A for text-to-video and video-to-text retrieval. COCO, VG, WV2M, CC3M are the abbreviations of COCO Caption~\cite{chen2015microsoft}, Visual Genome~\cite{krishna2017visual}, WebVid2M~\cite{bain2021frozen}, Google Conceptual Captions~\cite{sharma2018conceptual}, respectively.
  }
  \label{tab:MSRVTT}
\end{table*}

%% file: DiDeMo.tex
\begin{table}[!h]
  \centering
  \begin{tabular}{@{}l|cccc@{}}
    \multirow{2}*{\textbf{Method}}&\multicolumn{4}{c}{\textbf{Text $\Longrightarrow$ Video}}\\
     &  \textbf{R@1} & \textbf{R@5} & \textbf{R@10} & \textbf{MedR} \\
    \hline\hline
    S2VT~\cite{venugopalan2014translating}   &$11.9$   &$33.6$ &$-$ &$13.0$ \\
    FSE~\cite{zhang2018cross}    &$13.9$   &$36.0$ &$-$ &$11.0$ \\
    CE~\cite{liu2019use}     &$16.1$   &$44.1$ &$-$ &$8.3$ \\
    ClipBERT~\cite{lei2021less}  &$20.4$   &$44.5$ &$56.7$ &$7.0$\\
    Frozen~\cite{bain2021frozen}  &$31.0$   &$59.8$ &$\mathbf{72.4}$ &$3.0$ \\
    \hline
    \textbf{Ours}   &$\mathbf{32.5}$   &$\mathbf{60.8}$ &${72.3}$ &$\mathbf{3.0}$ \\
  \end{tabular}
  \caption{Comparisons with state-of-the-art results on DiDeMo for text-to-video retrieval.}
  \label{tab:DiDeMo}
\end{table}

%% file: LSMDC.tex
\begin{table}[!t]
  \centering
  \begin{tabular}{@{}l|llll@{}}
    \multirow{2}*{\textbf{Method}}&\multicolumn{4}{c}{\textbf{Text $\Longrightarrow$ Video}}\\
     &  \textbf{R@1} & \textbf{R@5} & \textbf{R@10} & \textbf{MedR} \\
    \hline\hline
    JSFusion~\cite{yu2018joint} &$9.1$ &$21.2$ &$34.1$ &$36.0$\\
    MEE~\cite{miech2018learning} &$9.3$ &$25.1$ &$33.4$ &$27.0$\\
    CE~\cite{liu2019use} &$11.2$ &$26.9$ &$34.8$ &$25.3$\\
    MMT~\cite{gabeur2020multi} &$12.9$ &$29.2$ &$38.8$ &$19.3$\\
    Frozen~\cite{bain2021frozen} &$15.0$ &$30.8$ &$39.8$ &$20.0$\\
    \hline
    \textbf{Ours} &$\mathbf{17.1}$ &$\mathbf{32.5}$ &$\mathbf{41.5}$ &$\mathbf{18.0}$\\
  \end{tabular}
  \caption{Comparisons with state-of-the-art results on LSMDC for text-to-video retrieval.}
  \label{tab:LSMDC}
\end{table}

%% file: MSVD.tex
\begin{table}[!t]
  \centering
  \begin{tabular}{@{}l|rrrr@{}}
    \multirow{2}*{\textbf{Method}}&\multicolumn{4}{c}{\textbf{Text $\Longrightarrow$ Video}}\\
     &  \textbf{R@1} & \textbf{R@5} & \textbf{R@10} & \textbf{MedR} \\
    \hline\hline
    VSE~\cite{kiros2014unifying}  &$12.3$ &$30.1$ &$42.3$ &$14.0$\\
    VSE++~\cite{faghri2017vse++} &$15.4$ &$39.6$ &$53.0$ &$9.0$\\
    Multi. Cues~\cite{mithun2018learning}  &$20.3$ &$47.8$ &$61.1$ &$6.0$\\
    CE~\cite{liu2019use} &$19.8$ &$49.0$ &$63.8$ &$6.0$\\
    Support Set~\cite{patrick2020support} &$23.0$ &$52.8$ &$65.8$ &$5.0$\\
    Support Set$^{\dagger}$~\cite{patrick2020support}  &$28.4$ &$60.0$ &$72.9$ &$4.0$\\
    Frozen~\cite{bain2021frozen}  &$33.7$ &$64.7$ &$76.3$ &$3.0$\\
    \hline
    \textbf{Ours}  &$\mathbf{44.0}$ &$\mathbf{74.9}$&$\mathbf{84.3}$&$\mathbf{2.0}$\\
  \end{tabular}
  \caption{Comparisons with state-of-the-art results on MSVD for text-to-video retrieval. $^{\dagger}$ indicates the method is pre-trained on HowTo100M.}
  \label{tab:MSVD}
\end{table}

%% file: component.tex
\begin{table}[!h]
  \centering
  \begin{tabular}{@{}l|llll@{}}
    \multirow{2}*{\textbf{Method}}&\multicolumn{4}{c}{\textbf{Text $\Longrightarrow$ Video}}\\
     &  \textbf{R@1} & \textbf{R@5} & \textbf{R@10} & \textbf{MedR} \\
    \hline\hline
    Region Learner &$\textbf{34.3}$ &$\textbf{60.2}$ &${72.0}$ &$\textbf{3.0}$\\
    - Interaction &$33.5$ &$60.4$ &$\textbf{72.5}$ &$\textbf{3.0}$ \\
    - Aggregation &$31.5$ &$\textbf{60.2}$ &${70.9}$ &$\textbf{3.0}$\\
    - Quantization&$29.4$ &$56.9$ &$69.0$ &$4.0$\\
  \end{tabular}
  \caption{Effectiveness of different component of our approach.}
  \label{tab:component}
\end{table}

%% file: tab6.tex
\begin{table}[!h]
  \centering
  \begin{tabular}{@{}l|llll@{}}
    \multirow{2}*{\textbf{Method}}&\multicolumn{4}{c}{\textbf{Text $\Longrightarrow$ Video}}\\
     &  \textbf{R@1} & \textbf{R@5} & \textbf{R@10} & \textbf{MedR} \\
    \hline\hline
    Base &$31.5$ &$\textbf{60.2}$ &${70.9}$ &$\textbf{3.0}$\\
    \hline
    + Random Sampling &$31.2$ &$60.3$ &$72.2$ &$\textbf{3.0}$\\
    + Manual Selection &$32.8$ &$59.7$ &$\textbf{72.4}$ &$\textbf{3.0}$\\
    + Naive Attention &$32.8$ &$59.6$ &$69.8$ &${4.0}$\\
    \hline
    + Region Mask &$\textbf{33.5}$ &$\textbf{60.4}$ &${72.5}$ &$\textbf{3.0}$\\
  \end{tabular}
  \caption{Effect of different feature aggregation.}
  \label{tab:num_regions}
\end{table}

%% file: region_depth.tex
\begin{figure}[h]
     \centering
     \begin{subfigure}[b]{0.236\textwidth}
         \centering
         \includegraphics[width=\textwidth]{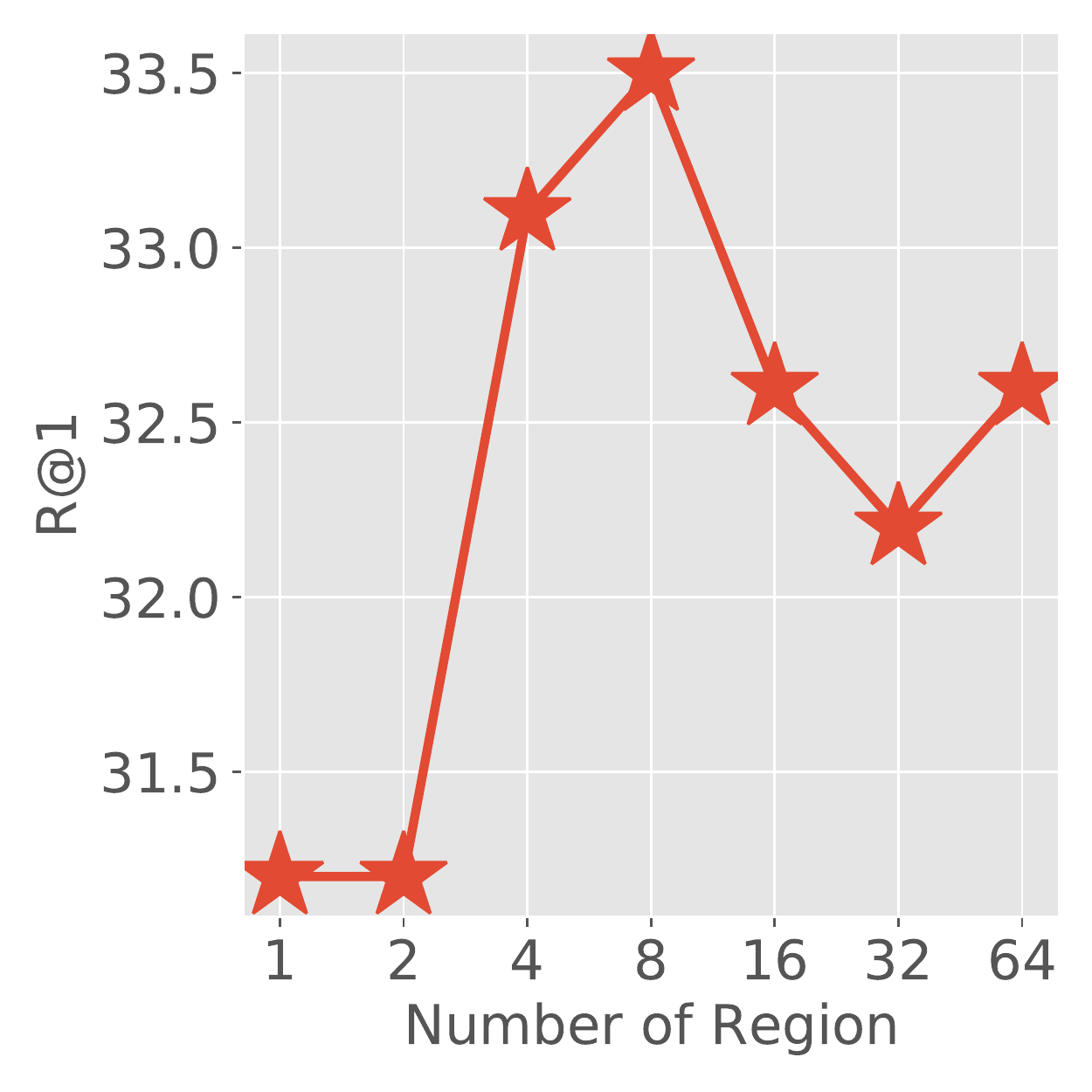}
         \caption{}
         \label{fig:num_region}
     \end{subfigure}
     \hfill
     \begin{subfigure}[b]{0.236\textwidth}
         \centering
         \includegraphics[width=\textwidth]{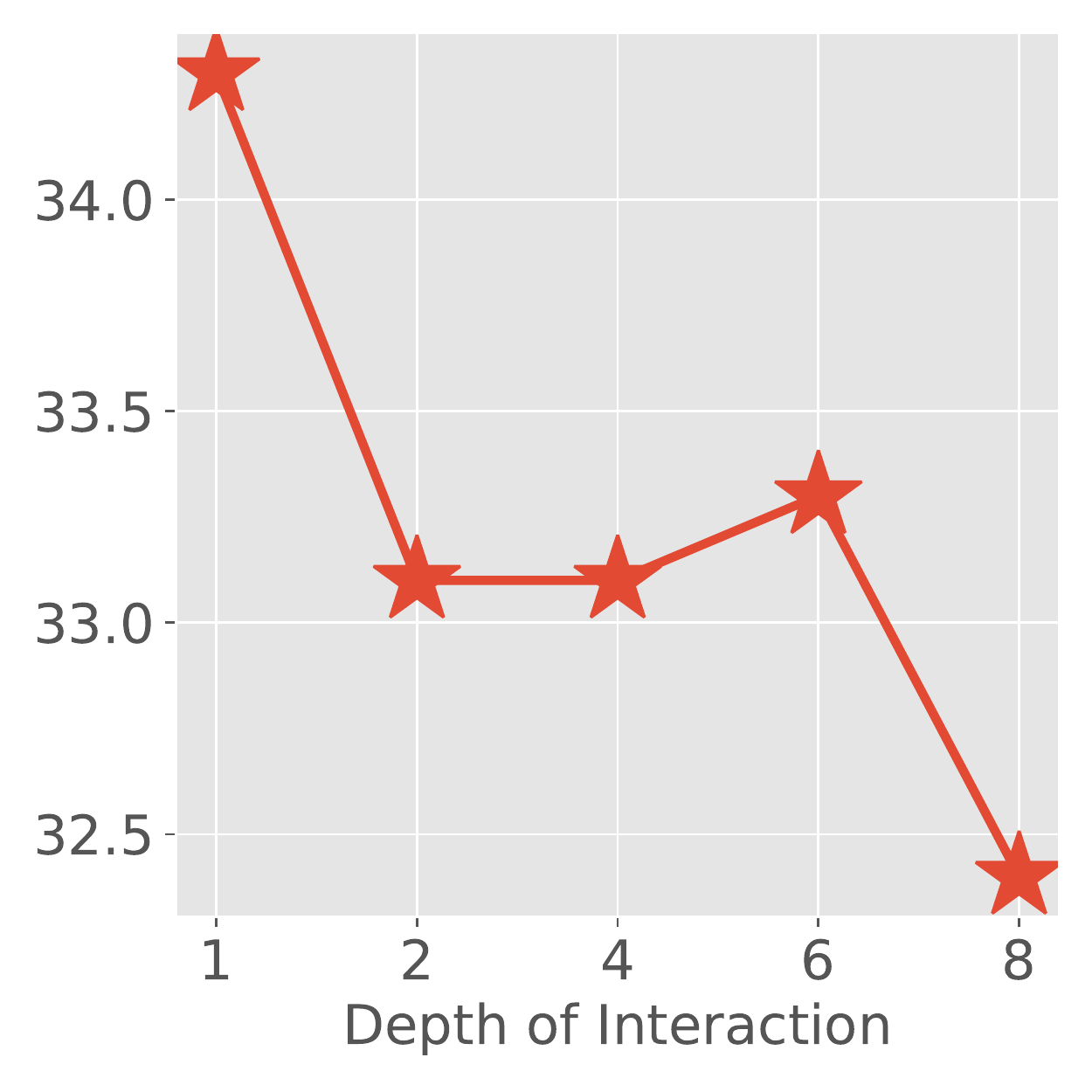}
         \caption{}
         \label{fig:depth}
     \end{subfigure}
        \caption{Effect of different number of regions and depth of interaction.}
\end{figure}

%% file: vis.tex
\begin{figure*}[t]
  \centering
  \includegraphics[width=1\linewidth]{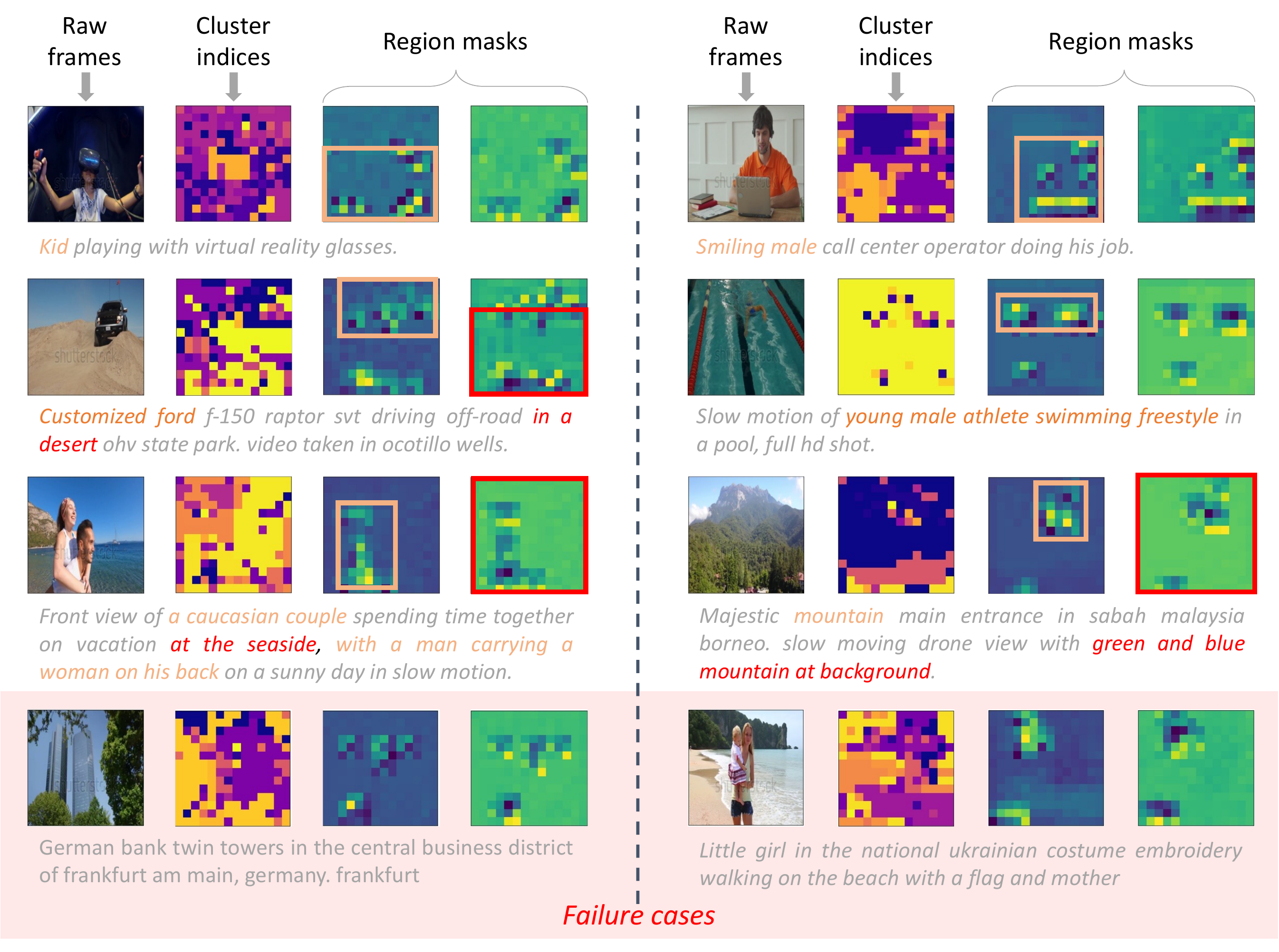}
   \caption{Visualization of the regions learned via the proposed Region Learner. Each group has a raw frame, the corresponding textual description, a learned map of cluster indices, and two selected learned region masks. We annotate the visual entities in colorful boxes for better understanding. The resolution of these learned maps is $14 \times 14$. (Best viewed in color.)}
   \label{fig:vis}
\end{figure*}

%% file: 5_conclusion.tex
\section{Conclusion and Limitations}
{
   To conclude, this work propose a simple yet effective RegionLearner to \textbf{quantize}, \textbf{aggregate}, and \textbf{interact} semantic features from visual objects/entities for better video-language alignment without any explicit supervision. It first quantize raw pixels into discrete latent embeddings, and then aggregate them into several regions via learnable masks, followed by interactions. The experimental results of four downstream benchmarks and some visualization results prove the effectiveness and interpretability of our method. We hope that RegionLearner can inspire future work for learning video-text representations with a more fine-grained way.

   \textbf{Limitations.}
   RegionLearner has the following potential limitations that would provide promising directions for future research, including i): exploring the robustness of RegionLearner on noisy video data, only part of which is aligned with the text description; ii): generalizing RegionLearner to the billion-scale pre-training datasets.

   
}

%% file: main.bbl
\begin{thebibliography}{10}\itemsep=-1pt

\bibitem{amrani2021noise}
Elad Amrani, Rami Ben-Ari, Daniel Rotman, and Alex Bronstein.
\newblock Noise estimation using density estimation for self-supervised
  multimodal learning.
\newblock In {\em AAAI}, pages 6644--6652, 2021.

\bibitem{anderson2018bottom}
Peter Anderson, Xiaodong He, Chris Buehler, Damien Teney, Mark Johnson, Stephen
  Gould, and Lei Zhang.
\newblock Bottom-up and top-down attention for image captioning and visual
  question answering.
\newblock In {\em CVPR}, pages 6077--6086, 2018.

\bibitem{anne2017localizing}
Lisa Anne~Hendricks, Oliver Wang, Eli Shechtman, Josef Sivic, Trevor Darrell,
  and Bryan Russell.
\newblock Localizing moments in video with natural language.
\newblock In {\em ICCV}, pages 5803--5812, 2017.

\bibitem{arandjelovic2016netvlad}
Relja Arandjelovic, Petr Gronat, Akihiko Torii, Tomas Pajdla, and Josef Sivic.
\newblock Netvlad: Cnn architecture for weakly supervised place recognition.
\newblock In {\em CVPR}, pages 5297--5307, 2016.

\bibitem{arnab2021vivit}
Anurag Arnab, Mostafa Dehghani, Georg Heigold, Chen Sun, Mario Lu{\v{c}}i{\'c},
  and Cordelia Schmid.
\newblock Vivit: A video vision transformer.
\newblock {\em arXiv preprint arXiv:2103.15691}, 2021.

\bibitem{bain2021frozen}
Max Bain, Arsha Nagrani, G{\"u}l Varol, and Andrew Zisserman.
\newblock Frozen in time: A joint video and image encoder for end-to-end
  retrieval.
\newblock {\em ICCV}, 2021.

\bibitem{bertasius2021space}
Gedas Bertasius, Heng Wang, and Lorenzo Torresani.
\newblock Is space-time attention all you need for video understanding?
\newblock {\em arXiv preprint arXiv:2102.05095}, 2021.

\bibitem{carreira2017quo}
Joao Carreira and Andrew Zisserman.
\newblock Quo vadis, action recognition? a new model and the kinetics dataset.
\newblock In {\em CVPR}, pages 6299--6308, 2017.

\bibitem{chen2011collecting}
David Chen and William~B Dolan.
\newblock Collecting highly parallel data for paraphrase evaluation.
\newblock In {\em ACL}, pages 190--200, 2011.

\bibitem{chen2015microsoft}
Xinlei Chen, Hao Fang, Tsung-Yi Lin, Ramakrishna Vedantam, Saurabh Gupta, Piotr
  Doll{\'a}r, and C~Lawrence Zitnick.
\newblock Microsoft coco captions: Data collection and evaluation server.
\newblock {\em arXiv preprint arXiv:1504.00325}, 2015.

\bibitem{chen2020uniter}
Yen-Chun Chen, Linjie Li, Licheng Yu, Ahmed~El Kholy, Faisal Ahmed, Zhe Gan, Yu
  Cheng, and Jingjing Liu.
\newblock {UNITER}: Universal image-text representation learning, 2020.

\bibitem{cheng2021motion}
Yi-Bin Cheng, Xipeng Chen, Dongyu Zhang, and Liang Lin.
\newblock Motion-transformer: self-supervised pre-training for skeleton-based
  action recognition.
\newblock In {\em ACM MM in Asia}, pages 1--6, 2021.

\bibitem{deng2009imagenet}
Jia Deng, Wei Dong, Richard Socher, Li-Jia Li, Kai Li, and Li Fei-Fei.
\newblock Imagenet: A large-scale hierarchical image database.
\newblock In {\em CVPR}, pages 248--255. Ieee, 2009.

\bibitem{devlin2018bert}
Jacob Devlin, Ming-Wei Chang, Kenton Lee, and Kristina Toutanova.
\newblock Bert: Pre-training of deep bidirectional transformers for language
  understanding.
\newblock {\em arXiv preprint arXiv:1810.04805}, 2018.

\bibitem{dosovitskiy2020image}
Alexey Dosovitskiy, Lucas Beyer, Alexander Kolesnikov, Dirk Weissenborn,
  Xiaohua Zhai, Thomas Unterthiner, Mostafa Dehghani, Matthias Minderer, Georg
  Heigold, Sylvain Gelly, et~al.
\newblock An image is worth 16x16 words: Transformers for image recognition at
  scale.
\newblock {\em arXiv preprint arXiv:2010.11929}, 2020.

\bibitem{faghri2017vse++}
Fartash Faghri, David~J Fleet, Jamie~Ryan Kiros, and Sanja Fidler.
\newblock Vse++: Improving visual-semantic embeddings with hard negatives.
\newblock {\em arXiv preprint arXiv:1707.05612}, 2017.

\bibitem{feichtenhofer2019slowfast}
Christoph Feichtenhofer, Haoqi Fan, Jitendra Malik, and Kaiming He.
\newblock Slowfast networks for video recognition.
\newblock In {\em ICCV}, pages 6202--6211, 2019.

\bibitem{feichtenhofer2016convolutional}
Christoph Feichtenhofer, Axel Pinz, and Andrew Zisserman.
\newblock Convolutional two-stream network fusion for video action recognition.
\newblock In {\em CVPR}, pages 1933--1941, 2016.

\bibitem{gabeur2020multi}
Valentin Gabeur, Chen Sun, Karteek Alahari, and Cordelia Schmid.
\newblock Multi-modal transformer for video retrieval.
\newblock In {\em ECCV}, pages 214--229, 2020.

\bibitem{gao2018ican}
Chen Gao, Yuliang Zou, and Jia-Bin Huang.
\newblock ican: Instance-centric attention network for human-object interaction
  detection.
\newblock {\em arXiv preprint arXiv:1808.10437}, 2018.

\bibitem{hara2018can}
Kensho Hara, Hirokatsu Kataoka, and Yutaka Satoh.
\newblock Can spatiotemporal 3d cnns retrace the history of 2d cnns and
  imagenet?
\newblock In {\em CVPR}, pages 6546--6555, 2018.

\bibitem{herzig2021object}
Roei Herzig, Elad Ben-Avraham, Karttikeya Mangalam, Amir Bar, Gal Chechik, Anna
  Rohrbach, Trevor Darrell, and Amir Globerson.
\newblock Object-region video transformers.
\newblock {\em arXiv preprint arXiv:2110.06915}, 2021.

\bibitem{huang2021seeing}
Zhicheng Huang, Zhaoyang Zeng, Yupan Huang, Bei Liu, Dongmei Fu, and Jianlong
  Fu.
\newblock Seeing out of the box: End-to-end pre-training for vision-language
  representation learning.
\newblock In {\em CVPR}, pages 12976--12985, 2021.

\bibitem{huang2020pixel}
Zhicheng Huang, Zhaoyang Zeng, Bei Liu, Dongmei Fu, and Jianlong Fu.
\newblock Pixel-bert: Aligning image pixels with text by deep multi-modal
  transformers.
\newblock {\em arXiv preprint arXiv:2004.00849}, 2020.

\bibitem{jang2017tgif}
Yunseok Jang, Yale Song, Youngjae Yu, Youngjin Kim, and Gunhee Kim.
\newblock Tgif-qa: Toward spatio-temporal reasoning in visual question
  answering.
\newblock In {\em CVPR}, pages 2758--2766, 2017.

\bibitem{jegou2010aggregating}
Herv{\'e} J{\'e}gou, Matthijs Douze, Cordelia Schmid, and Patrick P{\'e}rez.
\newblock Aggregating local descriptors into a compact image representation.
\newblock In {\em CVPR}, pages 3304--3311. IEEE, 2010.

\bibitem{karpathy2014large}
Andrej Karpathy, George Toderici, Sanketh Shetty, Thomas Leung, Rahul
  Sukthankar, and Li Fei-Fei.
\newblock Large-scale video classification with convolutional neural networks.
\newblock In {\em CVPR}, pages 1725--1732, 2014.

\bibitem{kay2017kinetics}
Will Kay, Joao Carreira, Karen Simonyan, Brian Zhang, Chloe Hillier, Sudheendra
  Vijayanarasimhan, Fabio Viola, Tim Green, Trevor Back, Paul Natsev, et~al.
\newblock The kinetics human action video dataset.
\newblock {\em arXiv preprint arXiv:1705.06950}, 2017.

\bibitem{kim2018bilinear}
Jin-Hwa Kim, Jaehyun Jun, and Byoung-Tak Zhang.
\newblock Bilinear attention networks.
\newblock In {\em NeurIPS}, pages 1564--1574, 2018.

\bibitem{kiros2014unifying}
Ryan Kiros, Ruslan Salakhutdinov, and Richard~S Zemel.
\newblock Unifying visual-semantic embeddings with multimodal neural language
  models.
\newblock {\em arXiv preprint arXiv:1411.2539}, 2014.

\bibitem{korbar2020video}
Bruno Korbar, Fabio Petroni, Rohit Girdhar, and Lorenzo Torresani.
\newblock Video understanding as machine translation.
\newblock {\em arXiv preprint arXiv:2006.07203}, 2020.

\bibitem{krishna2017dense}
Ranjay Krishna, Kenji Hata, Frederic Ren, Li Fei-Fei, and Juan Carlos~Niebles.
\newblock Dense-captioning events in videos.
\newblock In {\em ICCV}, pages 706--715, 2017.

\bibitem{krishna2017visual}
Ranjay Krishna, Yuke Zhu, Oliver Groth, Justin Johnson, Kenji Hata, Joshua
  Kravitz, Stephanie Chen, Yannis Kalantidis, Li-Jia Li, David~A Shamma, et~al.
\newblock Visual genome: Connecting language and vision using crowdsourced
  dense image annotations.
\newblock {\em IJCV}, 123(1):32--73, 2017.

\bibitem{lei2021less}
Jie Lei, Linjie Li, Luowei Zhou, Zhe Gan, Tamara~L Berg, Mohit Bansal, and
  Jingjing Liu.
\newblock Less is more: Clipbert for video-and-language learning via sparse
  sampling.
\newblock In {\em CVPR}, pages 7331--7341, 2021.

\bibitem{lei2018tvqa}
Jie Lei, Licheng Yu, Mohit Bansal, and Tamara~L Berg.
\newblock Tvqa: Localized, compositional video question answering.
\newblock {\em arXiv preprint arXiv:1809.01696}, 2018.

\bibitem{lei2019tvqa+}
Jie Lei, Licheng Yu, Tamara~L Berg, and Mohit Bansal.
\newblock Tvqa+: Spatio-temporal grounding for video question answering.
\newblock {\em arXiv preprint arXiv:1904.11574}, 2019.

\bibitem{li2020hero}
Linjie Li, Yen-Chun Chen, Yu Cheng, Zhe Gan, Licheng Yu, and Jingjing Liu.
\newblock Hero: Hierarchical encoder for video+ language omni-representation
  pre-training.
\newblock In {\em EMNLP}, 2020.

\bibitem{li2020oscar}
Xiujun Li, Xi Yin, Chunyuan Li, Pengchuan Zhang, Xiaowei Hu, Lei Zhang, Lijuan
  Wang, Houdong Hu, Li Dong, Furu Wei, et~al.
\newblock Oscar: Object-semantics aligned pre-training for vision-language
  tasks.
\newblock In {\em ECCV}, 2020.

\bibitem{lin2019tsm}
Ji Lin, Chuang Gan, and Song Han.
\newblock Tsm: Temporal shift module for efficient video understanding.
\newblock In {\em ICCV}, pages 7083--7093, 2019.

\bibitem{liu2021hit}
Song Liu, Haoqi Fan, Shengsheng Qian, Yiru Chen, Wenkui Ding, and Zhongyuan
  Wang.
\newblock Hit: Hierarchical transformer with momentum contrast for video-text
  retrieval.
\newblock {\em arXiv preprint arXiv:2103.15049}, 2021.

\bibitem{liu2019use}
Yang Liu, Samuel Albanie, Arsha Nagrani, and Andrew Zisserman.
\newblock Use what you have: Video retrieval using representations from
  collaborative experts.
\newblock In {\em BMVC}, 2019.

\bibitem{liu2021video}
Ze Liu, Jia Ning, Yue Cao, Yixuan Wei, Zheng Zhang, Stephen Lin, and Han Hu.
\newblock Video swin transformer.
\newblock {\em arXiv preprint arXiv:2106.13230}, 2021.

\bibitem{lu2019vilbert}
Jiasen Lu, Dhruv Batra, Devi Parikh, and Stefan Lee.
\newblock Vilbert: Pretraining task-agnostic visiolinguistic representations
  for vision-and-language tasks.
\newblock In {\em NeurIPS}, 2019.

\bibitem{luo2020univl}
Huaishao Luo, Lei Ji, Botian Shi, Haoyang Huang, Nan Duan, Tianrui Li, Jason
  Li, Taroon Bharti, and Ming Zhou.
\newblock Univl: A unified video and language pre-training model for multimodal
  understanding and generation.
\newblock {\em arXiv preprint arXiv:2002.06353}, 2020.

\bibitem{materzynska2020something}
Joanna Materzynska, Tete Xiao, Roei Herzig, Huijuan Xu, Xiaolong Wang, and
  Trevor Darrell.
\newblock Something-else: Compositional action recognition with
  spatial-temporal interaction networks.
\newblock In {\em CVPR}, pages 1049--1059, 2020.

\bibitem{miech2018learning}
Antoine Miech, Ivan Laptev, and Josef Sivic.
\newblock Learning a text-video embedding from incomplete and heterogeneous
  data.
\newblock {\em arXiv preprint arXiv:1804.02516}, 2018.

\bibitem{miech2019howto100m}
Antoine Miech, Dimitri Zhukov, Jean-Baptiste Alayrac, Makarand Tapaswi, Ivan
  Laptev, and Josef Sivic.
\newblock Howto100m: Learning a text-video embedding by watching hundred
  million narrated video clips.
\newblock In {\em ICCV}, pages 2630--2640, 2019.

\bibitem{mithun2018learning}
Niluthpol~Chowdhury Mithun, Juncheng Li, Florian Metze, and Amit~K
  Roy-Chowdhury.
\newblock Learning joint embedding with multimodal cues for cross-modal
  video-text retrieval.
\newblock In {\em ICMR}, pages 19--27, 2018.

\bibitem{neimark2021video}
Daniel Neimark, Omri Bar, Maya Zohar, and Dotan Asselmann.
\newblock Video transformer network.
\newblock {\em arXiv preprint arXiv:2102.00719}, 2021.

\bibitem{oord2017neural}
Aaron van~den Oord, Oriol Vinyals, and Koray Kavukcuoglu.
\newblock Neural discrete representation learning.
\newblock In {\em NeurIPS}, pages 6309--6318, 2017.

\bibitem{patrick2020support}
Mandela Patrick, Po-Yao Huang, Yuki Asano, Florian Metze, Alexander Hauptmann,
  Joao Henriques, and Andrea Vedaldi.
\newblock Support-set bottlenecks for video-text representation learning.
\newblock {\em arXiv preprint arXiv:2010.02824}, 2020.

\bibitem{portillo2021straightforward}
Jes{\'u}s~Andr{\'e}s Portillo-Quintero, Jos{\'e}~Carlos Ortiz-Bayliss, and Hugo
  Terashima-Mar{\'\i}n.
\newblock A straightforward framework for video retrieval using clip.
\newblock {\em arXiv preprint arXiv:2102.12443}, 2021.

\bibitem{radford2021learning}
Alec Radford, Jong~Wook Kim, Chris Hallacy, Aditya Ramesh, Gabriel Goh,
  Sandhini Agarwal, Girish Sastry, Amanda Askell, Pamela Mishkin, Jack Clark,
  et~al.
\newblock Learning transferable visual models from natural language
  supervision.
\newblock {\em arXiv preprint arXiv:2103.00020}, 2021.

\bibitem{redmon2016you}
Joseph Redmon, Santosh Divvala, Ross Girshick, and Ali Farhadi.
\newblock You only look once: Unified, real-time object detection.
\newblock In {\em CVPR}, pages 779--788, 2016.

\bibitem{renNIPS15fasterrcnn}
Shaoqing Ren, Kaiming He, Ross Girshick, and Jian Sun.
\newblock Faster {R-CNN}: Towards real-time object detection with region
  proposal networks.
\newblock In {\em NeurIPS}, 2015.

\bibitem{rohrbach2015dataset}
Anna Rohrbach, Marcus Rohrbach, Niket Tandon, and Bernt Schiele.
\newblock A dataset for movie description.
\newblock In {\em CVPR}, pages 3202--3212, 2015.

\bibitem{rohrbach2017movie}
Anna Rohrbach, Atousa Torabi, Marcus Rohrbach, Niket Tandon, Christopher Pal,
  Hugo Larochelle, Aaron Courville, and Bernt Schiele.
\newblock Movie description.
\newblock {\em IJCV}, 123(1):94--120, 2017.

\bibitem{rouditchenko2020avlnet}
Andrew Rouditchenko, Angie Boggust, David Harwath, Brian Chen, Dhiraj Joshi,
  Samuel Thomas, Kartik Audhkhasi, Hilde Kuehne, Rameswar Panda, Rogerio Feris,
  et~al.
\newblock Avlnet: Learning audio-visual language representations from
  instructional videos.
\newblock {\em arXiv preprint arXiv:2006.09199}, 2020.

\bibitem{ryoo2021tokenlearner}
Michael~S Ryoo, AJ Piergiovanni, Anurag Arnab, Mostafa Dehghani, and Anelia
  Angelova.
\newblock Tokenlearner: What can 8 learned tokens do for images and videos?
\newblock {\em arXiv preprint arXiv:2106.11297}, 2021.

\bibitem{sanh2019distilbert}
Victor Sanh, Lysandre Debut, Julien Chaumond, and Thomas Wolf.
\newblock Distilbert, a distilled version of bert: smaller, faster, cheaper and
  lighter.
\newblock {\em arXiv preprint arXiv:1910.01108}, 2019.

\bibitem{sharma2018conceptual}
Piyush Sharma, Nan Ding, Sebastian Goodman, and Radu Soricut.
\newblock Conceptual captions: A cleaned, hypernymed, image alt-text dataset
  for automatic image captioning.
\newblock In {\em ACL}, pages 2556--2565, 2018.

\bibitem{sigurdsson2016hollywood}
Gunnar~A Sigurdsson, G{\"u}l Varol, Xiaolong Wang, Ali Farhadi, Ivan Laptev,
  and Abhinav Gupta.
\newblock Hollywood in homes: Crowdsourcing data collection for activity
  understanding.
\newblock In {\em ECCV}, pages 510--526. Springer, 2016.

\bibitem{simonyan2014two}
Karen Simonyan and Andrew Zisserman.
\newblock Two-stream convolutional networks for action recognition in videos.
\newblock {\em arXiv preprint arXiv:1406.2199}, 2014.

\bibitem{Su2020VL-BERT}
Weijie Su, Xizhou Zhu, Yue Cao, Bin Li, Lewei Lu, Furu Wei, and Jifeng Dai.
\newblock Vl-bert: Pre-training of generic visual-linguistic representations.
\newblock In {\em ICLR}, 2020.

\bibitem{sun2019videobert}
Chen Sun, Austin Myers, Carl Vondrick, Kevin Murphy, and Cordelia Schmid.
\newblock Videobert: A joint model for video and language representation
  learning.
\newblock In {\em ICCV}, pages 7464--7473, 2019.

\bibitem{tan2019lxmert}
Hao Tan and Mohit Bansal.
\newblock Lxmert: Learning cross-modality encoder representations from
  transformers.
\newblock In {\em EMNLP}, pages 5103--5114, 2019.

\bibitem{tang2021decembert}
Zineng Tang, Jie Lei, and Mohit Bansal.
\newblock Decembert: Learning from noisy instructional videos via dense
  captions and entropy minimization.
\newblock In {\em NAACL}, pages 2415--2426, 2021.

\bibitem{tran2015learning}
Du Tran, Lubomir Bourdev, Rob Fergus, Lorenzo Torresani, and Manohar Paluri.
\newblock Learning spatiotemporal features with 3d convolutional networks.
\newblock In {\em ICCV}, pages 4489--4497, 2015.

\bibitem{vaswani2017attention}
Ashish Vaswani, Noam Shazeer, Niki Parmar, Jakob Uszkoreit, Llion Jones,
  Aidan~N Gomez, {\L}ukasz Kaiser, and Illia Polosukhin.
\newblock Attention is all you need.
\newblock In {\em NeurIPS}, pages 5998--6008, 2017.

\bibitem{venugopalan2014translating}
Subhashini Venugopalan, Huijuan Xu, Jeff Donahue, Marcus Rohrbach, Raymond
  Mooney, and Kate Saenko.
\newblock Translating videos to natural language using deep recurrent neural
  networks.
\newblock {\em arXiv preprint arXiv:1412.4729}, 2014.

\bibitem{wang2016temporal}
Limin Wang, Yuanjun Xiong, Zhe Wang, Yu Qiao, Dahua Lin, Xiaoou Tang, and Luc
  Van~Gool.
\newblock Temporal segment networks: Towards good practices for deep action
  recognition.
\newblock In {\em ECCV}, pages 20--36. Springer, 2016.

\bibitem{wang2018non}
Xiaolong Wang, Ross Girshick, Abhinav Gupta, and Kaiming He.
\newblock Non-local neural networks.
\newblock In {\em CVPR}, pages 7794--7803, 2018.

\bibitem{xu2016msr}
Jun Xu, Tao Mei, Ting Yao, and Yong Rui.
\newblock Msr-vtt: A large video description dataset for bridging video and
  language.
\newblock In {\em CVPR}, pages 5288--5296, 2016.

\bibitem{yan2020interactive}
Rui Yan, Lingxi Xie, Xiangbo Shu, and Jinhui Tang.
\newblock Interactive fusion of multi-level features for compositional activity
  recognition.
\newblock {\em arXiv preprint arXiv:2012.05689}, 2020.

\bibitem{yu2018joint}
Youngjae Yu, Jongseok Kim, and Gunhee Kim.
\newblock A joint sequence fusion model for video question answering and
  retrieval.
\newblock In {\em ECCV}, pages 471--487, 2018.

\bibitem{zhang2018cross}
Bowen Zhang, Hexiang Hu, and Fei Sha.
\newblock Cross-modal and hierarchical modeling of video and text.
\newblock In {\em ECCV}, pages 374--390, 2018.

\bibitem{zhang2021morphmlp}
David~Junhao Zhang, Kunchang Li, Yunpeng Chen, Yali Wang, Shashwat Chandra, Yu
  Qiao, Luoqi Liu, and Mike~Zheng Shou.
\newblock Morphmlp: A self-attention free, mlp-like backbone for image and
  video.
\newblock {\em arXiv preprint arXiv:2111.12527}, 2021.

\bibitem{zhou2018towards}
Luowei Zhou, Chenliang Xu, and Jason~J Corso.
\newblock Towards automatic learning of procedures from web instructional
  videos.
\newblock In {\em AAAI}, 2018.

\bibitem{zhu2020actbert}
Linchao Zhu and Yi Yang.
\newblock Actbert: Learning global-local video-text representations.
\newblock In {\em CVPR}, pages 8746--8755, 2020.

\bibitem{zhu2015aligning}
Yukun Zhu, Ryan Kiros, Rich Zemel, Ruslan Salakhutdinov, Raquel Urtasun,
  Antonio Torralba, and Sanja Fidler.
\newblock Aligning books and movies: Towards story-like visual explanations by
  watching movies and reading books.
\newblock In {\em ICCV}, pages 19--27, 2015.

\end{thebibliography}
